\documentclass[aps,prb,twocolumn,superscriptaddress]{revtex4-2}
\usepackage{algorithm}
\usepackage{algorithmic}
\usepackage{graphicx}
\usepackage{latexsym}
\usepackage{amssymb}
\usepackage{amsmath}
\usepackage{amsfonts}
\usepackage[vcentermath]{youngtab}
\usepackage{upgreek}
\usepackage{bm,dsfont}
\usepackage{multirow}
\usepackage{enumitem}
\usepackage{color}
\usepackage{hyperref}
\hypersetup{
colorlinks = true,
linkcolor = [rgb]{0.70,0.13,0.13},
citecolor = [rgb]{0.13,0.55,0.13},
urlcolor = [rgb]{0.25, 0.41, 0.88}}
\usepackage{makecell}

\newcommand{\bra}[1]{\langle #1|}
\newcommand{\ket}[1]{|#1 \rangle}
\newcommand{\braket}[2]{\langle #1|#2 \rangle}

\newcommand{\dsN}{\mathbb{N}}

\newcommand{\vect}[1]{{\bm{#1}}}

\newcommand{\eqnref}[1]{Eq.\,\eqref{#1}}
\newcommand{\figref}[1]{Fig.\,\ref{#1}}

\newcommand{\secref}[1]{Sec.\,\ref{#1}}

\newcommand{\bea}{\begin{eqnarray}}
\newcommand{\eea}{\end{eqnarray}}
\def\be{\begin{equation}}
\def\ee{\end{equation}}
\newcommand{\beq}{\begin{equation}}
\newcommand{\eeq}{\end{equation}}
\newcommand{\beqn}{\begin{eqnarray}}
\newcommand{\eeqn}{\end{eqnarray}}

\begin{document}

\title{Sequential learning on a Tensor Network Born machine with Trainable Token Embedding}

\author{Wanda Hou}
\affiliation{Department of Physics, University of California, San Diego, CA 92093, USA}
\author{Miao Li}
\affiliation{School of Physics, Zhejiang University, Hangzhou 310027, China}
\affiliation{Department of Physics, Georgia Institute of Technology,
Atlanta, GA 30332, USA}
\author{Yi-Zhuang You}
\thanks{Corresponding author: yzyou@physics.ucsd.edu}
\affiliation{Department of Physics, University of California, San Diego, CA 92093, USA}
\date{\today}

\begin{abstract}
 Generative models aim to learn the probability distributions underlying data, enabling the generation of new, realistic samples. Quantum-inspired generative models, such as Born machines based on the matrix product state (MPS) framework, have demonstrated remarkable capabilities in unsupervised learning tasks. This study advances the Born machine paradigm by introducing trainable token embeddings through positive operator-valued measurements (POVMs), replacing the traditional approach of static tensor indices. Key technical innovations include encoding tokens as quantum measurement operators with trainable parameters and leveraging QR decomposition to adjust the physical dimensions of the MPS. This approach maximizes the utilization of operator space and enhances the model’s expressiveness. Empirical results on RNA data demonstrate that the proposed method significantly reduces negative log-likelihood (NLL) compared to one-hot embeddings, with higher physical dimensions further enhancing single-site probabilities and multi-site correlations. The model also outperforms GPT-2 in single-site estimation and achieves competitive correlation modeling, showcasing the potential of trainable POVM embeddings for complex data correlations in quantum-inspired sequence modeling.
\end{abstract}

\maketitle

\section{Introduction}

Unsupervised learning is crucial for leveraging the vast amount of unlabeled data available today. One important approach is generative modeling, which aims to model the probability distribution of data given a presumed underlying physical mechanism\cite{lecun2015deep,2019RvMP...91d5002C}. Different from the discriminative models that aim to find the decision boundary of different classes, the goal of the generative model is to replicate the underlying distribution $p(x)$ of observed data, and then generate new samples that are plausible within the context of the original dataset. 

Generative modeling techniques have made rapid advancements in recent years, many models, such as Variational Auto Encoders (VAEs)\cite{2013arXiv1312.6114K}, Generative Adversarial Networks (GANs)\cite{2014arXiv1406.2661G,2018PhRvA..98a2324D}, and Transformers\cite{2017arXiv170603762V}, have been applied across various domains. Drawing inspiration from classical physics, present models have more or less employed the idea of energy-based modeling, where certain probabilities are modeled as $p(x)\propto e^{-E(x)}$ following Boltzmann distribution as in statistical physics\cite{ackley1985learning,salakhutdinov2015learning}. In contrast, quantum physics employs wave function models, where probabilities are modeled as $p(x)=|\Psi(x)|^2$, with $\Psi(x)$ being a quantum wave function. This class of quantum generative models is known as the Born machines\cite{PhysRevX.8.031012, 2022PhRvA.106b2612K, 2021arXiv211205326M}. Quantum probability models are more expressive than their classical counterparts, as they can model distributions that violate Bell’s inequality, a characteristic of quantum entanglement\cite{trueblood2012quantum,haven2016quantum}. 

The Born machines have many featured advantages as being a generative modeling algorithm, such as tractable log-likelihood and marginalized log-likelihood, which provides better interpretability and enables anomaly detection; efficient and exact evaluation of the log-likelihood gradient enables sampling-free gradient-based optimization; supporting both autoregressive and masked sampling. Due to these features, the Born machines have shown excellent performance in many domains of unsupervised learning tasks such as sequential data learning, and can outperform models based on classical physics\cite{2018Entrp..20..583C,2018arXiv180404168L}.

A particularly important class of data is the one-dimensional sequential data, which refers to a set of data points that are ordered based on a specific sequence, typically governed by a temporal or spatial order. Born machines are well-developed for sequence modeling due to the one-dimensional structure of sequences that allows for a highly efficient tensor network representation, called the matrix product state (MPS), for modeling the underlying quantum state\cite{orus2019tensor,2019PhRvX...9c1041L,2021arXiv210612974L}. Born machines support tractable log-likelihood that provides better interpretability and efficient direct sampling, supporting both autoregressive and masked sampling. The current state of research has made tremendous progress by leveraging the entanglement feature of quantum states and encoding them into an MPS to improve the expressiveness of Born machines\cite{2018PhRvA..98a2324D,mitarai2018quantum,2018arXiv180404168L,gong2022born,2019arXiv190300556M}.

For modeling discrete types of sequential data $\vect{x}=(x_1,x_2,\cdots)$, the corresponding tokens $x_i\in \dsN$ are directly treated as the tensor index of the physical dimension of MPS, and encoded as a diagonal basis projector $\ket{x_i}\bra{x_i}$. However, in this study, we explore a more general embedding method such that encoding each token $x_i$ as a more general quantum measurement operator $M_{\gamma}(x_i)$ in the local Hilbert space with trainable parameters $\gamma$ rather than as a diagonal basis projector. The probabilities are modeled as $p_{\theta,\gamma}(\vect{x})=\bra{\Psi_\theta}M_{\gamma}(\vect{x})\ket{\Psi_\theta}$, where $M_\gamma(\vect{x})=\bigotimes_i M_\gamma(x_i)$ is the joint measurement operator corresponding to the sequence $\vect{x}$, and $\ket{\Psi_\theta}$ is a many-body quantum state, still structured as an MPS parametrized by $\theta$. The key innovation is that, encoding the token as a measurement operator fully utilizes the operator space, allowing for the packing of more tokens in a much smaller Hilbert space. Furthermore, by increasing the physical dimension of both embedding quantum measurements and MPS, the model can utilize the power of larger operator space to improve expressiveness. Our experimental result on sequential RNA data shows that, with the flexibility of adjusting physical dimensions, the Born machines can exhibit better performance and learn deeper correlations from the dataset.

\section{Encoding structure}

The goal of generative modeling is to model the joint probability $p(\vect{x})$ given the training dataset. Once the model is well-trained, it can be used to generate new samples that should ideally resemble the distribution of the training data. In this work, we adopt the traditional Born Machine structure, which uses quantum wave function $\ket{\Psi_\theta}$ as the backbone and encoding tokens as quantum measurement operators $M(\vect{x})=\{M_{\gamma}(x_i)|x_i\in V\}$, where $\gamma$, $\theta$ are trainable parameters from embedding and MPS respectively, and $V$ is the total vocabulary set. The joint probability of each data point $\vect{x}=(x_1,x_2,...,x_n)$ can be embedded by the expectation value of quantum measure $M_\gamma(\vect{x})=\bigotimes_{i}M_{\gamma}(x_i)$ given quantum wave function.
\begin{equation}\label{eq: P}
    p_{\theta, \gamma}(\vect{x})=\bra{\Psi_\theta}\bigotimes_{i}M_{\gamma}(x_i)\ket{\Psi_\theta},
\end{equation}
where the backbone quantum state is normalized $\braket{\Psi_\theta}{\Psi_\theta}=1$. In order to guarantee that the encoded probabilities are \textit{real}, \textit{positive semi-definite} and \textit{normalize} with respect to the entire input space, we restrict the embedded operators to be positive operator-valued measurements (POVMs), which are \textit{hermitian}, \textit{positive semi-definite} and overall \textit{normalized}:
\begin{equation}\label{eq: emb}
\begin{split}
    \forall x_{i}\in V,\ M_{\gamma}(x_i)&=M^{\dagger}_{\gamma}(x_i)\succeq 0,\\
    \sum_{i}M_{\gamma}(x_i)&=I.
\end{split}
\end{equation}
where $I$ stands for the identity operator. Previous research has explored various implementations of adaptive POVMs \cite{2022arXiv220807817G, 2023arXiv230315353Z, 2021PRXQ....2d0342G}; this work extends this concept by employing QR decomposition for trainable token embeddings in sequential data modeling in Born machines.

In the previous Born machine model parametrization, the embedding constraint is satisfied by assigning each token $x_i\in\dsN$ directly by a corresponding projector $\ket{x_i}\bra{x_i}$ of the basis state\cite{PhysRevX.8.031012} labeled by the token index. This is equivalent to fixing the POVMs as one-hot embedding matrices $M_\text{one-hot}(x_i)_{\alpha\beta}=\delta_{\alpha,x_i}\delta_{\beta,x_i}$ with the matrix dimension strictly equals to the vocabulary size. Although the one-hot embedding could satisfy \eqnref{eq: emb}, this embedding scheme is predefined and not trainable. It cannot fully utilize the operator space and is suboptimal. In contrast, the trainable POVM embedding allows for potentially packing more tokens in the same Hilbert space, and further putting the physical dimension of the quantum state as an additional hyper-parameter to improve the expressive power of the model. In this paper, we propose a trainable parametrization method that can leverage the potential of the operator space while satisfying \eqnref{eq: emb}.

The parameters $\gamma$ are initialized as a complex tensor of the shape $(v \times p, p)$, where $v$ stands for vocabulary size and $p$ stands for the physical dimension of the quantum state. Then $\gamma$ is forwarded with a QR decomposition $\gamma=Q_\gamma R_\gamma$ and the resulting $Q_{\gamma}$ matrix is further reshaped into $(v, p, p)$ then batch-multiplied with its hermitian conjugate to get the quantum measurement matrices. 
\begin{equation}\label{eq: parametrization}
    M_{\gamma}(x_{i})=Q^{\dagger}_{\gamma}(x_i)Q_{\gamma}(x_i), \ x_i\in V.
\end{equation}
The above construction ensures that the constraints in \eqnref{eq: emb} are automatically satisfied and the physical dimension becomes an adjustable hyper-parameter.

\begin{figure}[htbp]
\begin{center}
\includegraphics[width=240pt]{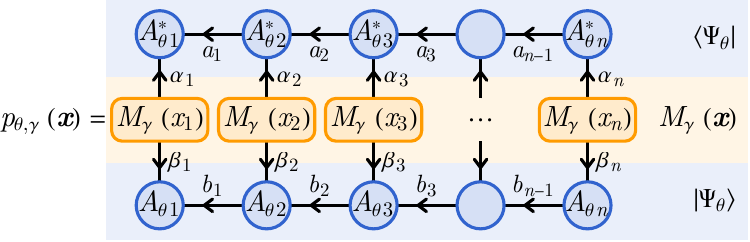}
\caption{Tensor network representation of probability model $p_{\theta,\gamma}(\vect{x})$. The MPS tensors $A_{\theta i}$ are blue circles, and the measurement operators $M_\gamma(x_i)$ are orange boxes. The isometric structure is indicated by arrows on the tensor legs, projecting from the larger Hilbert space to the smaller sub-space.}
\label{fig: tn}
\end{center}
\end{figure}

Following \cite{PhysRevX.8.031012}, the backbone quantum state can be efficiently parametrized with normalized MPS, and the corresponding isometric constrain $\braket{\Psi_\theta}{\Psi_\theta}=1$ is maintained by isometric optimization method during gradient-based optimization\cite{wen2013feasible,2022MLS&T...3a5020G}. With an explicit choice of basis $\ket{\vect{\beta}}=\ket{\beta_1,\beta_2,\cdots,\beta_n}$ in the physical Hilbert space, the MPS wave function can be expressed as 
\begin{equation}\label{eq: mps}
\braket{\vect{\beta}}{\Psi_\theta}=\sum_{b_1 b_2...b_{n-1}}(A_{\theta 1})^{\beta_1}_{b_1} (A_{\theta 2})^{\beta_2}_{b_1 b_2}\dots (A_{\theta n})^{\beta_n}_{b_{n-1}},
\end{equation}
which are modeled by a series of tensors $A_{\theta i}$ parameterized by $\theta$ at site $i$. The tensors $A_{\theta i}$ are isometric in the sense that after contracting the in-coming legs between $A_{\theta i}$ and $A_{\theta i}^*$ (complex conjugation), the out-going legs from an identity map,
\begin{equation}
\sum_{b_i,\beta_i} (A_{\theta i}^*)_{b_{i-1}b_{i}}^{\beta_i}(A_{\theta i})_{b'_{i-1}b_{i}}^{\beta_i}=\delta_{b_{i-1}b'_{i-1}.}
\end{equation}
The isometry structure is indicated in \figref{fig: tn} by the arrows. Employing the isometric MPS has several advantages. First, it ensures the normalization of the modeled probability distribution, allowing direct log-likelihood optimization. Second, 
it allows efficient marginalization of the probability distribution to the leading token, by summing out the subsequent tokens.

Given the tractable log-likelihood of the Born machine, the unsupervised learning objective function for a specified dataset is the negative log-likelihood (NLL) \begin{equation}
\mathcal{L}=-\frac{1}{N}\sum_{\vect{x}\in \text{data}}\log p_{\theta,\gamma}(\vect{x}),
\end{equation}
where $N$ denotes the size of the dataset. \figref{fig: tn} shows the tensor network of the probability encoding of a single data point in one-dimensional shape $\vect{x}=(x_1,x_2,...,x_n)$. The training process for optimizing the parameters based on the NLL is outlined in Algorithm~\ref{alg:train}.

\begin{algorithm}[H]
\caption{Training Process of the Born Machine}
\label{alg:train}
\begin{algorithmic}
\REQUIRE Training dataset $D$, learning rate $\eta$
\STATE Initialize parameters $\theta$, $\gamma$
\WHILE{not converged}
    \FOR{each data sample $x \in D$}
        \STATE Compute joint probability $p_{\theta,\gamma}(x)$ using \eqnref{eq: P}
        \STATE Compute NLL loss $L = -\log p_{\theta,\gamma}(x)$
        \STATE Compute gradients $\nabla_{\theta} L$, $\nabla_{\gamma} L$
        \STATE Update parameters:
        \STATE $\theta \leftarrow \theta - \eta \nabla_{\theta} L$
        \STATE $\gamma \leftarrow \gamma - \eta \nabla_{\gamma} L$
    \ENDFOR
    \STATE Check for convergence
\ENDWHILE
\ENSURE Optimized parameters $\theta$, $\gamma$
\end{algorithmic}
\end{algorithm}

\section{Marginalization and Generative sampling}

After training, the Born machine enables autoregressive generation according to \eqnref{eq: P}, while in contrast, the energy-based model typically requires Markov chain Monte Carlo (MCMC) to sample from $p(\vect{x})=\frac{1}{Z}e^{-E(\vect{x})}$. The Born machine's autoregressive sampling procedure runs bit by bit, relying on its capability of conditional/marginalized probabilities evaluation. Importantly, the sampling direction is not restricted from the beginning or in any direction due to the fact that tensor contraction operation is commutative. Therefore the sites with given tokens can be conditioned by inserting the corresponding POVMs and then traced out, while the stand-by sites can be marginalized by directly tracing out the corresponding physical legs of MPS, benefiting from the normalized property in \eqnref{eq: emb}. The generation of a whole sequence is completed by iteratively implementing the same procedure to fill in all the vacant sites.

For example, let $K\subseteq\{1,2,...,n\}$ be a subset of sites, and $\vect{x}_{K}=\{x_i\}_{i\in K}$ be the tokens within the region $K$. After training, suppose we want to evaluate the conditional distribution $p_{\theta,\gamma}(\vect{x}_K|\vect{x}_{\bar{K}})$, given the remaining tokens $\vect{x}_{\bar{K}}$ in the complement region $\bar{K}$. It can be evaluated as
\begin{equation}\label{eq: single site}
    p_{\theta,\gamma}(\vect{x}_K|\vect{x}_{\bar{K}})=\bra{\Psi_\theta(\vect{x}_{\bar{K}})}M_{\gamma}(\vect{x}_K)\ket{\Psi_\theta(\vect{x}_{\bar{K}})},
\end{equation}
where $M_{\gamma}(\vect{x}_K)=\bigotimes_{i\in K}M_{\gamma}(x_i)$ is the measurement operator within region $K$, and $\ket{\Psi_\theta(\vect{x}_{\bar{K}})}$ denotes the collapsed quantum state according to the measurement outcomes $\vect{x}_{\bar{K}}$ in the region $\bar{K}$, defined as
\begin{equation}
\ket{\Psi_\theta(\vect{x}_{\bar{K}})}=\frac{1}{Z}\bigotimes_{i\in\bar{K}}Q_{\gamma}(x_i)\ket{\Psi_{\theta}},
\end{equation}
with $Z=\bra{\Psi_\theta}M(\vect{x}_{\bar{K}})\ket{\Psi_\theta}$ being the state normalization constant.
Unlike conventional autoregressive models, which only allow the conditional probability distribution to be evaluated following the causal order, the Born machine here enables the evaluation of the probability distribution of tokens in any subset $K$ conditioned on those in the complement set $\bar{K}$. 

\begin{algorithm}[H]
\caption{Generative Sampling with Conditional Dependencies}
\label{alg:gen}
\begin{algorithmic}
\REQUIRE Trained parameters $\theta$, $\gamma$, sequence length $n$, optional initial sequence $x_{\text{init}}$
\IF{$x_{\text{init}}$ is not provided}
    \STATE Initialize sequence $x = [ \text{empty} ]$ of length $n$
\ELSE
    \STATE Use $x_{\text{init}}$ as the starting sequence, where elements can be empty or predefined tokens
\ENDIF
\STATE Define set $K$ as indices of known tokens and $\bar{K}$ as indices of tokens to be sampled
\WHILE{$\bar{K}$ is not empty}
    \FOR{each index $i$ in $\bar{K}$}
        \IF{$x[i]$ is empty}
            \STATE Compute conditional probabilities for token $x_i$ given the known tokens using \eqnref{eq: single site}:
            \STATE $p(x_i|x_{K}) = \langle \Psi_{\theta}(x_{K}) | M_{\gamma}(x_i) | \Psi_{\theta}(x_{K}) \rangle$
            \STATE Sample $x_i$ from $p(x_i|x_{K})$
            \STATE Update $x[i]$ with sampled token $x_i$
            \STATE Add $i$ to set $K$ and remove $i$ from $\bar{K}$
        \ENDIF
    \ENDFOR
\ENDWHILE
\ENSURE Generated sequence $x$
\end{algorithmic}
\end{algorithm}

Since the Born machine enables autoregressive generation in arbitrary order, one can leverage this feature to do direct masked sampling or anomaly detection by evaluating the single site probabilities \eqnref{eq: single site} or joint probabilities of multiple sites. Moreover, one can compare the marginalized probabilities learned by the model with the statistical frequencies from the training dataset to evaluate the performance of trained models, which will be further discussed in \secref{Empirical result}. The detailed implementation of this process is outlined in Algorithm~\ref{alg:gen}.

\section{Empirical result}\label{Empirical result}

Due to the one-dimensional structure of MPS, Born machines are particularly well-suited for sequence modeling. Among various types of sequential data, the biological gene sequence is one of the most vital and holds immense value for humans. However, it is challenging to obtain experimentally and often incurs significant costs to extract limited data. Therefore, using generative models to discern the underlying probability distribution in gene sequences and generate new samples can enhance the efficiency of exploring new gene sequences and reduce costs. In this study, we will evaluate the performance of Born machines with trainable POVMs for the unsupervised learning of RNA sequences. We uses the bactaria's 5S rRNA with average length 120 and total vocabulary size of four(\textit{A}: Adenine, \textit{T}: Thymine, \textit{C}: Cytosine, \textit{G}: Guanine) from open source database \textit{5SrRNAdb}\cite{szymanski20165srnadb}. The training set and test set are unbiasedly divided from the dataset.

\begin{figure}[htbp]
\begin{center}
\includegraphics[width=250pt]{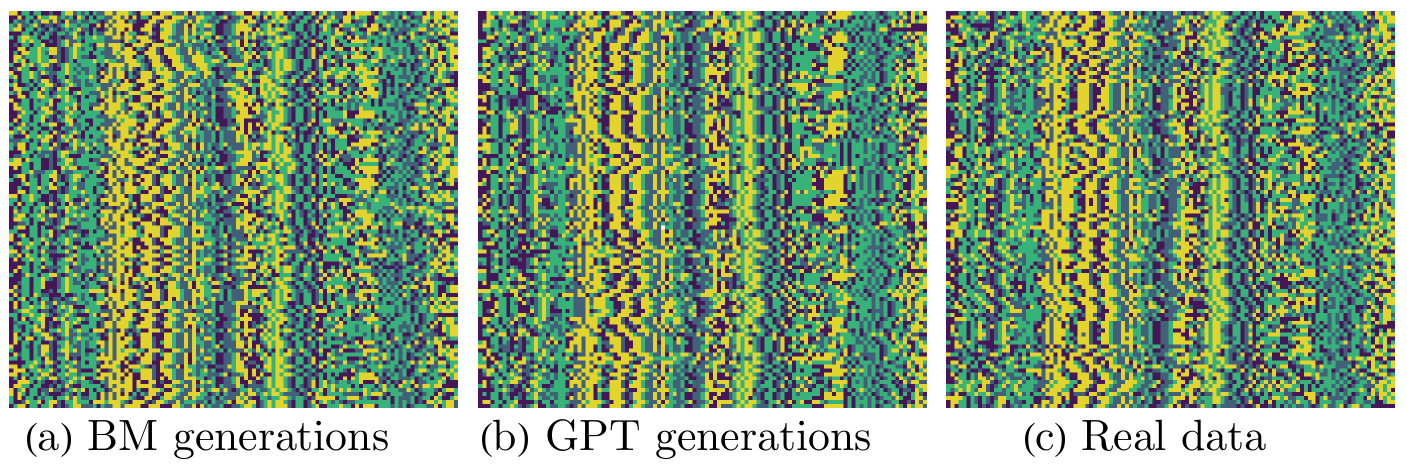}
\caption{In this figure, each horizontal line represents an RNA sequence, displayed vertically. The four colors represent four types of nucleotides. (a) Generated samples by Born machine with POVM embedding. (b) Generated samples by GPT-2. (c) Real data from the test set.}
\label{fig: generation}
\end{center}
\end{figure}

\begin{figure}[htbp]
\begin{center}
\includegraphics[width=200pt]{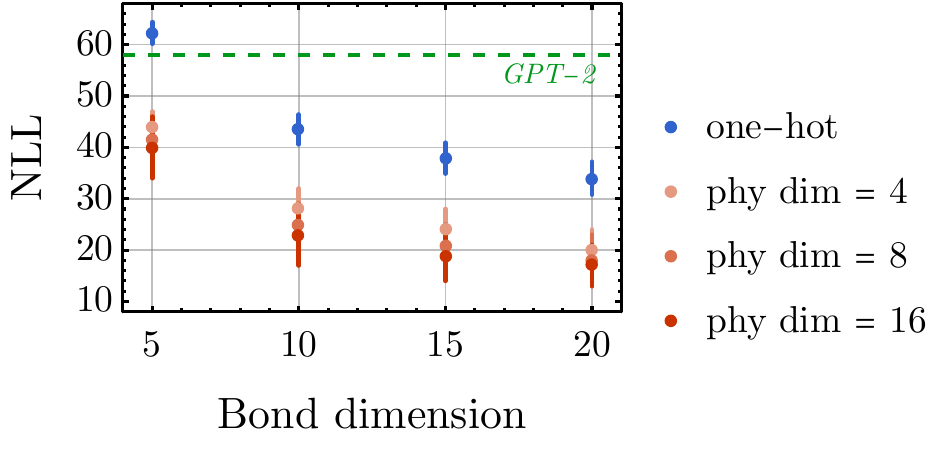}
\caption{The NLL objective function is minimized using \textit{Adam} optimizer with adjustable learning rate for bond/physical dimension separately. Result shows that the converged NLL is decreased by jointly increasing the bond dimension and the physical dimension. The green dashed line represents the converged NLL of GPT-2 model with similar parameter size.}
\label{fig: loss}
\end{center}
\end{figure}

\figref{fig: generation} generated by the Born machine with POVMs and GPT-2 appear qualitatively similar to the real test set, indicating both models capture visible patterns in the sequence data. Given this similarity, we proceed to evaluate each model’s performance using quantitative metrics such as NLL, single-site probabilities, and Pearson correlations. \figref{fig: loss} demonstrates that increasing the physical dimension in POVMs reduces the converged NLL values, even with a fixed boundary dimension. Furthermore, it reveals that higher physical dimensions extend the upper limit of bond dimension, enabling further NLL minimization.

\begin{figure}[htbp]
\begin{center}
\includegraphics[width=240pt]{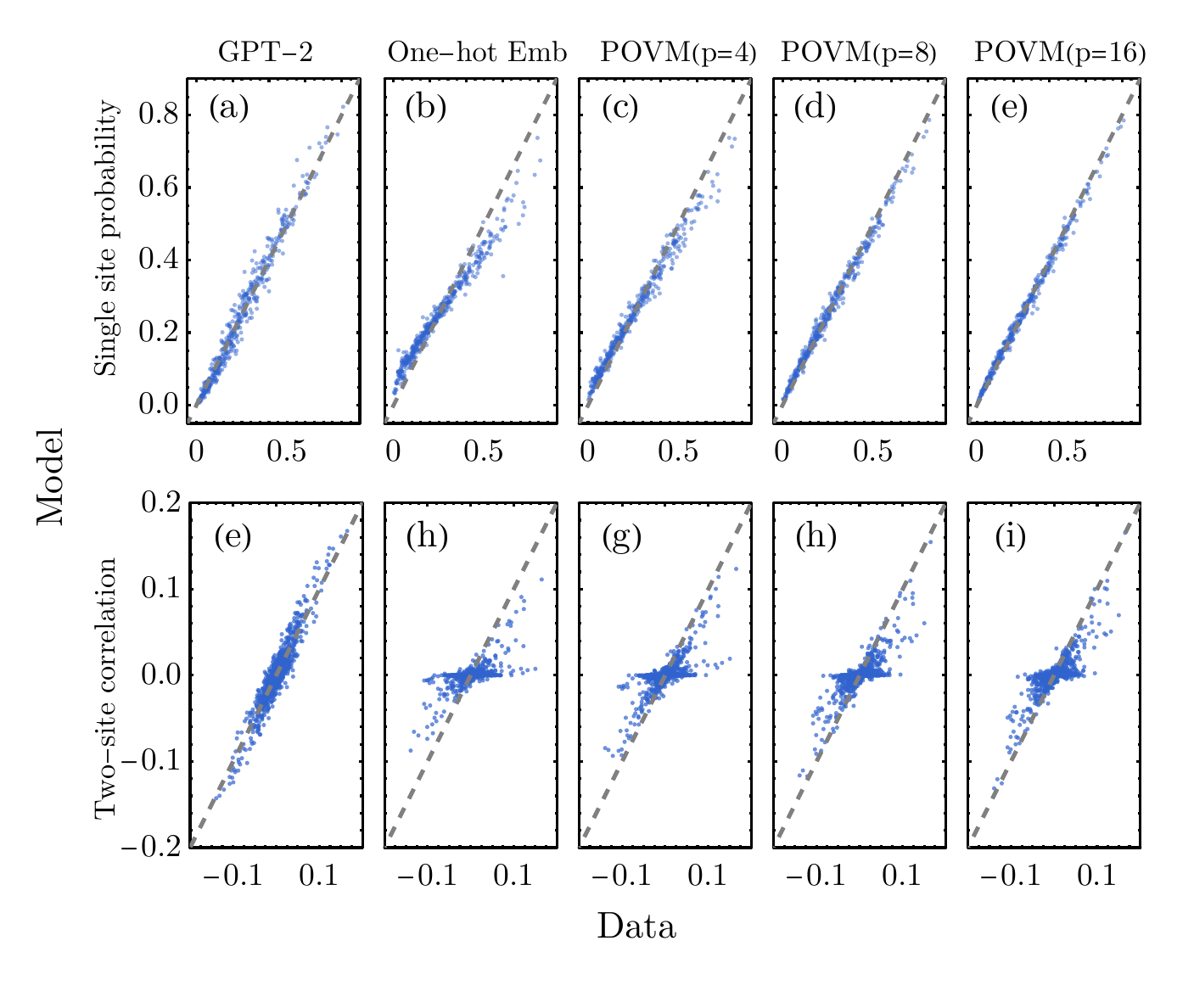}
\caption{Compare (a-e) the single-site probability $p(x_i)$ (upper row) and (f-j) the two-site correlation $c(x_i,y_j)$ (lower row) of the model with those from the dataset. GPT-2: (a)\&(f). One-hot embedding (baseline): (b)\&(g). Trainable POVM embedding (ours) with the physical dimension: (c)\&(h) $p=4$, (d)\&(i) $p=8$, (e)\&(j) $p=16$. The dashed line represents the $y=x$ reference line.}
\label{fig: corre}
\end{center}
\end{figure}

One of the features of RNA data distribution is that, the marginalized joint probability distribution among various sites reveals the underlying contact prediction information\cite{trinquier2021efficient}. Contact prediction in RNA refers to the process of determining which residues or nucleotides in an RNA molecule are in close proximity to each other in the molecule's three-dimensional structure. To delve deeper into evaluating the performance of the Born machine with trainable POVMs and discovering contact prediction, one can leverage its efficient marginalization ability \eqnref{eq: single site} and then compare it to statistical frequencies extracted from the dataset. In particular, we will study two \textit{statistical features}:
\begin{itemize}
\item the single-site probability distribution $p(x_i)$,
\item and the two-site Pearson correlation
\begin{equation}\label{eq: corre}
    c(x_i,x_j)=\log\frac{p(x_i,x_j)}{p(x_i)p(x_j)}.
\end{equation}
\end{itemize}
Each sub-figure in \figref{fig: corre} plots the model-predicted statistical feature (vertical axis) against the data-indicated statistical feature (horizontal axis). \figref{fig: corre}(a-e) is for $p(x_i)$ where each point in the plot corresponds to a different choice of site $i$ and a token $x_i$ on that site. \figref{fig: corre}(f-j) is for $c(x_i,x_j)$ where each point in the plot corresponds to a choice of pair $(i,j)$ and tokens $(x_i,x_j)$ on that pair. If the model learns all statistical features from data perfectly, we should expect all points to fall along the diagonal dashed line (on which the model prediction coincides with the data statistics).  The result in \figref{fig: corre} indicates that as the physical dimension increases, both the single-site probability and correlations align more closely with the test set. This indicates that the Born machine equipped with trainable POVM embedding will be more powerful in modeling token correlations across different sites in the training data, compared to the fixed one-hot embedding approach.

We also benchmarked our quantum model against the classical sequential generative model, the Generative Pretrained Transformer-2 (GPT-2), with a comparable parameter size. The GPT-2 configuration is initialized from the \textit{Hugging Face transformer} library\cite{gpt2lmhead} with $d_{\text{embedding}} = 72$, $n_{\text{layer}} = 12$, $n_{\text{head}} = 4$, which contains approximately 0.8 million parameters in total, and is integrated with the GPT-2LMHead model. For the Born machine, the total number of parameters can be estimated as the product of the bond dimension, physical dimension, and sequence length, i.e., $d_{\text{bond}}^2 \times d_{\text{phys}} \times L$. For example, our largest configuration uses $d_{\text{bond}} = 20$, $d_{\text{phys}} = 16$, and $L = 120$, leading to approximately 0.77 million parameters—comparable to GPT-2. In \figref{fig: loss} and \figref{fig: corre}, the Born machine achieves lower NLL and comparable performance in approximating single-site probabilities relative to GPT-2 under our current settings. However, it shows weaker proficiency in correlation approximation. Attention-based networks like GPT-2 are designed with an all-to-all connectivity structure without inherent dimensionality constraints, making them particularly effective at capturing long-range correlations.

\section{Summary}

Generative models are a vital branch of machine learning that strive to understand the inherent probability distribution of data. Unlike discriminative models, they focus on capturing the data's intrinsic structure. This work continues on studying quantum-inspired model called the Born machines for generative modeling. These machines use the quantum wave function to model underlying data distribution over the matrix product state (MPS) framework. The main innovation in this study is the extension of the token embedding method. Instead of embedding each token directly by a corresponding tensor index, this work introduces the concept of trainable positive operator-valued measurements (POVMs). By combining MPS with trainable embeddings, Born machines can achieve better performance and extract deeper correlations from datasets. The source code and raw data are available in the GitHub repository\cite{hou2023github}.

Our new Born machine architecture leverages quantum wave functions as its backbone, with tokens encoded as quantum measurement matrices. The joint probability of each data point is expressed as the expectation value of a quantum measurement. A key feature of the Born machine is its capability for autoregressive generation, enabling sequence sampling in arbitrary order—making it versatile for tasks such as masked token sampling and anomaly detection. While our current implementation is classically simulable via matrix product states (MPS), the true advantage may arise when sampling is carried out on a real quantum device. A promising strategy is to train the model efficiently on classical hardware using MPS, and then transfer the trained state to a quantum computer via an equivalent quantum circuit. As demonstrated in \cite{2022PhRvX..12a1047H, 2024QS&T....9a5012R}, each MPS tensor can be mapped to local quantum gates with ancilla insertions and measurements in a quantum circuit, representing the MPS as a real quantum state on the quantum computer. Furthermore, the adaptive POVM method introduced in this work can be implemented via generalized measurements on an extended Hilbert space \cite{preskill1998lecture}, enabling efficient sampling and potentially leveraging quantum advantage. Since POVMs can be directly realized in practice, this approach can be extended to the Quantum Circuit Born Machine (QCBM) architecture for implementation on quantum hardware \cite{PhysRevA.98.062324, 2019npjQI...5...45B}.

In terms of application, our study focuses on the unsupervised learning of RNA sequences. Results indicate that by increasing the physical dimension in the POVMs, the Born machine can achieve better performance, capturing deeper correlations in the RNA data. However, there are still unresolved issues within the current framework of the Born machine. For instance, its present structure cannot directly handle data with significant length variances. Moreover, because the current Born machine is constructed using a one-dimensional MPS with a discretized embedding method, it's awkward to directly extend it to handle higher-dimensional and continuous data types. Additionally, there's the challenge of transferring the well-trained MPS and corresponding POVMs to a quantum device for efficient sampling.

\begin{acknowledgments}
We acknowledge the helpful discussions with Rose Yu, Tian-Xiao Hu and Zhao-Yi Zeng. 
We acknowledge the OpenAI GPT4 model for providing editing suggestions throughout the process of writing this paper. The research is supported by the NSF Grant No.~DMR-2238360.
\end{acknowledgments}

\bibliographystyle{apsrev4-2}
\bibliography{ref}

\onecolumngrid
\newpage
\appendix
\setcounter{secnumdepth}{2}

\end{document}